\documentclass[sigconf]{acmart}
\AtBeginDocument{%
  }

\copyrightyear{2026}
\acmYear{2026}
\setcopyright{cc}
\setcctype{by}
\acmConference[MM '26]{Proceedings of the 34th ACM International Conference on Multimedia}{November 10--14, 2026}{Rio de Janeiro, Brazil}
\acmBooktitle{Proceedings of the 34th ACM International Conference on Multimedia (MM '26), November 10--14, 2026, Rio de Janeiro, Brazil}
\acmDOI{10.1145/3767308.3835372}
\acmISBN{979-8-4007-2213-4/2026/11}

\usepackage{tabularx}
\usepackage{multirow}
\usepackage{multicol}
\usepackage{tcolorbox}
\usepackage{subfig,graphicx}
\usepackage{fontawesome}
\usepackage{pifont}
\usepackage{graphicx}
\usepackage{listings}
\usepackage{fontawesome}
\usepackage{tcolorbox}
\tcbuselibrary{most} 
\definecolor{boxframe}{RGB}{40, 60, 85} 
\definecolor{boxback}{RGB}{248, 250, 252} 
\definecolor{boxtitle}{RGB}{230, 236, 242}

\newtcolorbox{greybox}{
    colback=gray!10,
    colframe=black,
    arc=1.2mm,
    boxrule=0.5pt,
    boxsep=0pt,
    left=4pt,
    right=4pt,
    top=1pt,
    bottom=3pt,
    boxsep=0pt,      
    before skip=5pt,   
    after skip=5pt,   
}

\usepackage{multicol}
\usepackage{multirow}
\usepackage{booktabs,array,graphicx,tikz}
\usepackage[table]{xcolor}
\usepackage{colortbl}         % colors
\usepackage{tabularx, booktabs}

\begin{document}

\title{M3MAD-Bench: Multi-Dimensional Evaluation  of \\ Multi-Agent Debate Across Domains and Modalities}


\author{Ao Li}
\authornote{These authors contributed equally to this work.}
\affiliation{%
  \institution{Shandong University}
  \city{Qingdao}
  \country{China}}
\email{liaolea@mail.sdu.edu.cn}

\author{Jinghui Zhang}
\authornotemark[1]
\affiliation{%
  \institution{Mohamed bin Zayed University of Artificial Intelligence}
  \city{Abu Dhabi}
  \country{UAE}}
\email{jinghui.zhang@mbzuai.ac.ae}

\author{Luyu Li}
\affiliation{%
  \institution{Shandong University}
  \city{Qingdao}
  \country{China}}
\email{LuyuLi@mail.sdu.edu.cn}

\author{Yuxiang Duan}
\affiliation{%
  \institution{Shandong University}
  \city{Qingdao}
  \country{China}}
\email{yuxiangd@mail.sdu.edu.cn}

\author{Lang Gao}
\affiliation{%
  \institution{Mohamed bin Zayed University of Artificial Intelligence}
  \city{Abu Dhabi}
  \country{UAE}}
\email{lang.gao@mbzuai.ac.ae}

\author{MingCai Chen}
\affiliation{%
  \institution{Nanjing University of Posts and Telecommunications}
  \city{Nanjing}
  \country{China}}
\email{chenmc@njupt.edu.cn}

\author{Weijun Qin}
\affiliation{%
  \institution{EBTech Co. Ltd.}
  \city{Beijing}
  \country{China}}
\email{qinweijun99@gmail.com}

\author{Shaopeng Li}
\affiliation{%
  \institution{EBTech Co. Ltd.}
  \city{Beijing}
  \country{China}}
\email{shaopengl@ebtech.com}

\author{Fengxian Ji}
\affiliation{%
  \institution{Mohamed bin Zayed University of Artificial Intelligence}
  \city{Abu Dhabi}
  \country{UAE}}
\email{fengxian.ji@mbzuai.ac.ae}

\author{Ning Liu}
\affiliation{%
  \institution{Shandong University}
  \city{Jinan}
  \country{China}}
\email{liun21cs@sdu.edu.cn}

\author{Lizhen Cui}
\affiliation{%
  \institution{Shandong University}
  \city{Jinan}
  \country{China}}
\email{clz@sdu.edu.cn}

\author{Xiuying Chen}
\affiliation{%
  \institution{Mohamed bin Zayed University of Artificial Intelligence}
  \city{Abu Dhabi}
  \country{UAE}}
\email{xiuying.chen@mbzuai.ac.ae}

\author{Yuntao Du}
\correspondingauthor
\affiliation{%
  \institution{Shandong University}
  \city{Jinan}
  \country{China}}
\email{yuntaodu@sdu.edu.cn}

\renewcommand{\shortauthors}{Li et al.}

\begin{abstract}
As an agent-level reasoning and coordination paradigm, Multi-Agent Debate (MAD) orchestrates multiple agents through structured debate to improve answer quality and support complex reasoning. 
However, existing research on MAD suffers from two fundamental limitations: evaluations are conducted under fragmented and inconsistent settings, making fair comparison difficult, and are largely confined to text-only scenarios, leaving its effectiveness in multimodal settings underexplored.
To address these gaps, we introduce \textbf{M3MAD-Bench}, a unified and extensible benchmark for evaluating MAD methods across \textbf{M}ulti-domain tasks, \textbf{M}ulti-modal inputs, and \textbf{M}ulti-dimensional metrics. 
M3MAD-Bench establishes standardized protocols over five core task domains, including Knowledge, Mathematics, Medicine, Natural Sciences, and Complex Reasoning, covering a total of 13 datasets, and systematically includes both pure text and vision-language data, enabling controlled cross-modality comparison.
We evaluate MAD methods on 9 base models spanning different architectures, scales, and modality capabilities.
Beyond accuracy, M3MAD-Bench incorporates efficiency-oriented metrics such as token consumption and inference time, providing a holistic view of performance--cost trade-offs. 
Through extensive experiments, we derive nine key insights, revealing that MAD is not uniformly effective: collaborative methods are generally more robust than adversarial ones, especially on reasoning-intensive and multimodal tasks, but often incur substantial efficiency costs.
These findings provide practical guidance for selecting and designing MAD strategies in real-world applications.
We believe M3MAD-Bench offers a reliable foundation for future research on standardized and reproducible MAD evaluation. 
The code is available at ~\url{https://github.com/liaolea/M3MAD-Bench}.
\end{abstract}





\maketitle

\section{Introduction}
\label{sec:intro}

\begin{figure*}[t]
\includegraphics[width=1\linewidth]{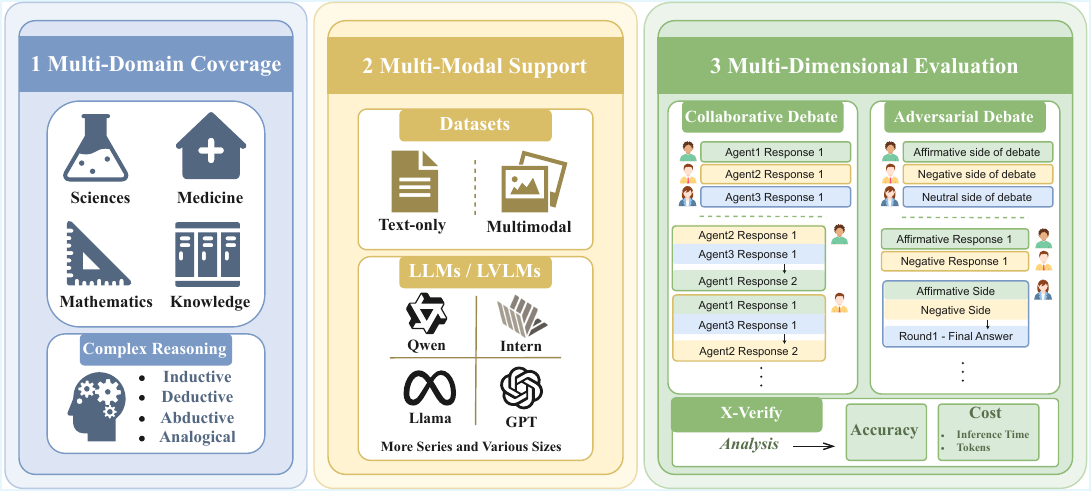} 
\caption{Overview of M3MAD-Bench. The framework comprises three pillars: (1) \textit{Multi-Domain Coverage} spanning diverse subjects and complex reasoning; (2) \textit{Multi-Modal Support} integrating both text-only and vision-language datasets; and (3) \textit{Multi-Dimensional Evaluation} assessing Collaborative and Adversarial debate strategies via accuracy and cost metrics.}
\label{fig:main}
\end{figure*}

The rapid advancement of Large Language Models (LLMs) has driven the development of LLM-based agents as a promising paradigm for further improving model performance~\cite{huang2024understanding,guo2024large,zhang1}. 
By orchestrating complex reasoning and decision-making processes, these agent systems have demonstrated strong capabilities across a wide range of tasks.
Among existing multi-agent approaches, \emph{Multi-Agent Debate} (MAD) has emerged as one of the most widely adopted and empirically effective strategies~\cite{liu2025truth,tian2025symbolic}. 
In MAD, multiple agents engage in structured debates, exchanging arguments and counterarguments to collaboratively reason toward more accurate conclusions.
Its appeal lies in a simple yet powerful intuition: interaction among multiple agents can expose errors, aggregate complementary evidence, and improve reasoning reliability beyond what a single model can achieve.

Despite its demonstrated promise, current studies on MAD suffer from several fundamental limitations that hinder a systematic understanding of its behavior and performance.
First, prior studies are conducted on \textit{fragmented and inconsistent setups}, including dataset disparities, base model heterogeneity, and non-uniform task configurations.~\cite{debate2,debate1}. 
Such inconsistencies make it difficult to fairly compare methods or isolate the actual contribution of debate strategies. 
Second, existing MAD methods are predominantly evaluated under \textit{unimodal, text-only settings}~\cite{zhu2025multiagentbench}, which severely restricts their applicability to real-world scenarios characterized by rich multimodal information.
As a result, the effectiveness of MAD in multimodal environments remains largely unexplored.

To address these challenges, we introduce \textbf{M3MAD-Bench}, a unified and extensible benchmark for the systematic evaluation of MAD methods.
As illustrated by Figure~\ref{fig:main}, M3MAD-Bench is designed around three core ``multi'' dimensions.
First, it provides \textit{Multi-Domain coverage} spanning five major domains, specifically knowledge, mathematics, medicine, natural sciences, and complex reasoning, which comprises a total of 13 representative  datasets split into 7 text-only and 6 multimodal datasets, enabling evaluation beyond a narrow set of commonly reused tasks and supports more comprehensive conclusions about MAD behavior. 
Second, it extends \textit{Multi-Modal evaluation} beyond text-only scenarios by incorporating vision--language inputs and systematically evaluating base models with diverse architectures, scales, and modality capabilities, enabling controlled and fair cross-modality comparison.
Third, it offers \textit{multi-dimensional metrics} that go beyond accuracy to include cost-related measurements, namely token consumption and inference time, thereby providing a more practical view of performance--cost trade-offs.

Built upon this comprehensive benchmark design, we conduct a systematic assessment of both the effectiveness and efficiency of MAD methods across text-only and multimodal settings, yielding nine in-depth insights: (1) MAD methods show mixed effectiveness: some approaches can improve over single-agent baselines, while others may underperform; (2) collaborative debate consistently outperforms adversarial paradigms in both performance and robustness across all modalities; (3) the effectiveness of MAD varies across tasks, with more pronounced improvements on logic- and reasoning-oriented benchmarks than on knowledge-focused ones; (4) introducing multiple perspectives in debate improves multimodal performance but offers little or negative gains in unimodal tasks; (5) MAD effectiveness is affected by the performance of the underlying base model; (6) MAD methods incur high token and time costs with limited performance gains; (7) the inherent differences among heterogeneous models do not necessarily translate into distinct reasoning paths, and in most cases, do not lead to performance improvements; (8) more debate rounds offer little benefit, while more agents generally improve performance; and (9) agents tend to reinforce mutual errors rather than self-correcting, and aggregators often fail to identify valid reasoning already present in the debate history.
In addition, we release a unified open-source evaluation codebase to ensure reproducibility and facilitate future research on MAD methods.

In summary, our contributions are as follows: 

% \textbf{Systematic Review:} 
\noindent $\bullet$ We identify two critical gaps in MAD research: the lack of standardized evaluation protocols across fragmented studies and the limited exploration of MAD's effectiveness in multimodal scenarios.

% \textbf{M3MAD-Bench \& Codebase:} 
\noindent $\bullet$ We introduce \textbf{M3MAD-Bench}, a unified benchmark for MAD across Multi-domain tasks (13 datasets), Multi-modal inputs, and Multi-dimensional metrics. To support reproducibility, we provide an open-source codebase that offers standardized implementations for debate orchestration and efficiency-oriented measurement.

% \textbf{Extensive Evaluation:} 
\noindent $\bullet$ Through systematic experiments, we derive nine in-depth insights. We demonstrate that MAD is not uniformly effective: while collaborative methods excel in reasoning and multimodal tasks, they often face challenges such as agent "stubbornness," high token costs, and the risk of reinforcing mutual errors (collective delusion).

\section{Related Work}

\noindent \textbf{Large Language Models and Multimodal Extensions.} Large Language Models (LLMs)~\cite{llama3,qwen25} have demonstrated remarkable capabilities in language understanding, reasoning, and generation. Benefiting from large-scale pretraining and instruction tuning, they have become the foundation for a wide range of modern language-based intelligent systems. Building on this progress, Large Vision-Language Models (LVLMs)~\cite{qwen25vl,internvl3,gpt-4o-mini,video-xl-pro,emoverse} extend LLMs to multimodal settings by integrating visual perception with linguistic reasoning. Recent LVLMs typically adopt a tightly coupled architecture that combines a large language backbone, a vision encoder for perceptual representation~\cite{dosovitskiy2020image}, and explicit cross-modal alignment modules to connect heterogeneous modalities.

\noindent \textbf{Multi-agent Debate.}
Multi-Agent Debate (MAD) is a reasoning framework where multiple agents iteratively exchange feedback and refine their answers through discussion. This paradigm has been applied to a variety of tasks and has shown potential for a wide range of tasks such as improving reasoning quality and factual accuracy~\cite{bo2024reflective, debate1, chanchateval, tang2024medagents, wuautogen, chenagentverse}. To further enhance MAD performance, some studies have incorporated debate-theory-inspired improvements~\cite{xiong2023examining}, while others proposed mechanisms such as Peer Rank and Peer Discussion to select suitable agent pairs for interaction~\cite{liprd}.
Considerable work has also focused on optimizing communication structures and protocols to boost the efficiency and effectiveness of debates~\cite{chanchateval,liu2024groupdebate,liu2024dynamic,li2024improving,phamlet,zhang2024cut}. In addition, ensuring diversity among agents has been highlighted as a key factor, whether by employing heterogeneous architectures~\cite{chen2024reconcile}, assigning distinct personas to each agent~\cite{liu2024dynamic,debate2,wang2024unleashing}, or controlling the diversity of generated text~\cite{liu2025breaking, chu2024exploring}. Finally, learning-based approaches have been explored to adaptively optimize MAD dynamics and improve overall debate outcomes~\cite{liu2024dynamic, estornellacc,chen2024optima}.

\noindent \textbf{Multi-Agent Debate Benchmark.}
Despite growing interest in MAD, the field still lacks a standardized evaluation protocol. Existing studies \citep{debate1,debate2} typically evaluate ad-hoc subsets of reasoning tasks (e.g., MATH, MMLU) under inconsistent hyperparameter settings (e.g., agent number, temperature), hindering reproducible comparison across debate frameworks. Although recent benchmarks attempt to standardize multi-agent evaluation, they fail to capture debate-specific dynamics. General benchmarks such as MASLab \citep{maslab} and MultiAgentBench \citep{zhu2025multiagentbench} are largely limited to text-only reasoning or execution tasks, and rely on limited metrics, typically accuracy and token consumption,  overlooking other critical aspects, such as inference latency. M3MAD-Bench addresses these limitations by a unified benchmark tailored to MAD.

\section{M3MAD-Bench}
In this section, we present the design of M3MAD-Bench, covering the evaluated methods, datasets, and metrics.

\subsection{Multi-Agent Debate Methods} 
Given the high sensitivity of multi-agent performance to prompt variations \citep{sclar2024quantifying}, we avoid re-implementing closed-source methods to ensure reproducibility. Instead, we select three representative MAD approaches with diverse interaction dynamics and evaluate them using their official public implementations. As illustrated in Figure~\ref{fig:main}(3), these methods fall into two paradigms: \textit{Collaborative Debate} and \textit{Adversarial Debate}.

\textit{Collaborative Debate} seeks consensus through information sharing and peer review among agents. We consider two representative methods: \textbf{LLM Debate} \citep{debate1}, which models standard consensus, and \textbf{DMAD} \citep{liu2025breaking}, which enhances collaboration via diversity. LLM Debate, inspired by the “Society of Minds” hypothesis, relies on iterative discussion and redundancy to reduce stochastic hallucinations. To address the risk of reinforcing shared errors, DMAD introduces \textit{Diversity of Thought}, encouraging distinct reasoning paths and enabling a comparison between strategic diversity and simple redundancy \citep{li2024improving}.

In contrast to these collaborative frameworks, \textit{Adversarial Debate} paradigm is designed to address the inherent risk of ``groupthink''. 
% where agents blindly conform to an incorrect majority.
The representative method is \citep{debate2}, which we refer to as \textbf{Div-MAD} to distinguish it from the general MAD acronym.
Div-MAD adopts a conflict-driven interaction through a ``Tit-for-Tat'' strategy, explicitly compelling one agent to act as a ``Devil's Advocate'' to challenge another.

As shown in Table~\ref{tab:maslab_comparison}, besides debate-based methods, M3MAD-Bench includes several representative baselines: standard Input–Out\allowbreak put prompting (\textbf{IO}), \textbf{Chain-of-Thought (CoT)} \cite{wei2022chain}, and \textbf{Self-Consistency (SC)} \cite{sc}. These strong non-debate baselines allow us to isolate the benefits of multi-agent interaction from those of enhanced single-model reasoning. 
By evaluating all methods under identical backbone models and datasets, M3MAD-Bench provides a controlled comparison to determine whether and when multi-agent debate yields more reliable multimodal decision-making, beyond gains from longer reasoning traces or multiple sampling.

\begin{table}[t]
\centering
\small
\renewcommand{\arraystretch}{1.2} 
\resizebox{\linewidth}{!}{%
\begin{tabular}{l c c c}
\toprule
\raisebox{1ex}{\textbf{Methodology}} & \textbf{\shortstack{Flexible\\Agent Numbers}} & \textbf{\shortstack{Peer\\Interaction}} & \textbf{\shortstack{Iterative\\Refinement}} \\
\midrule
\rowcolor{gray!15}\multicolumn{4}{l}{\textit{Single-Agent Baselines}} \\
IO &- & -& -\\
CoT~\cite{wei2022chain} & -& -& -\\
\midrule
\rowcolor{gray!15}\multicolumn{4}{l}{\textit{Multi-Agent Baseline}} \\
SC~\cite{sc} & \textcolor{green!50!black}{$\checkmark$} & \textcolor{red}{\ding{55}} & \textcolor{red}{\ding{55}}\\
\midrule
\rowcolor{gray!15}\multicolumn{4}{l}{\textit{Multi-Agent Debate Frameworks}} \\
LLM Debate~\cite{debate1} & \textcolor{green!50!black}{$\checkmark$} & \textcolor{green!50!black}{$\checkmark$} & \textcolor{green!50!black}{$\checkmark$} \\
Div-MAD~\cite{debate2} & \textcolor{red}{\ding{55}} & \textcolor{green!50!black}{$\checkmark$}& \textcolor{green!50!black}{$\checkmark$} \\
DMAD~\cite{liu2025breaking} & \textcolor{green!50!black}{$\checkmark$} & \textcolor{green!50!black}{$\checkmark$} & \textcolor{green!50!black}{$\checkmark$} \\
\bottomrule
\end{tabular}%
}
\caption{Overview of the methods supported in M3MAD-Bench. Flexible Agent Numbers indicates whether the number of agents or samples can be scaled, Peer Interaction denotes explicit information exchange among agents, and Iterative Refinement refers to multi-round optimization over intermediate responses.}
\label{tab:maslab_comparison}
% \vspace{-2mm}
\end{table}

\subsection{Datasets}

To comprehensively evaluate the effectiveness of MAD methods across diverse domains, we define five commonly used capability dimensions and manually curate a total of 13 representative datasets covering both unimodal and multimodal settings.

For single-modal evaluation, we cover Knowledge represented by MMLU~\cite{mmlu} and MMLU-Pro~\cite{mmlu-pro}, Mathematics including MATH~\cite{math} and GSM-Hard~\cite{gsm-hard}, Medicine featuring MedMCQA~\cite{medmcqa} and MedQA \cite{medqa}, and Natural Sciences represented by GPQA~\cite{gpqa}. 
For multi-modal evaluation, we include Knowledge from MME \cite{mme}, Mathematics via MathVista \cite{mathvista} and MathVision \cite{mathvision}, Medicine using PathVQA \cite{pathvqa}, and Reasoning through MME-Reasoning \cite{mme-reasoning} and VisualPuzzles \cite{visualpuzzles}, where the reasoning capability is comprehensively defined to encompass inductive, deductive, abductive, and analogical reasoning, all of which are covered by the selected datasets.

This domain- and modality-diverse selection enables a comprehensive assessment of MAD methods under both unimodal and multimodal settings, capturing a broad range of real-world reasoning and understanding challenges.

\begin{table*}[t]
    \centering
    \small
    \setlength{\tabcolsep}{5.2pt}
    \renewcommand{\arraystretch}{1.05}
    \begin{tabular*}{0.9\textwidth}{@{\extracolsep{\fill}}llcccccccc@{}}
    \toprule
        \multirow{2}{*}{Model} & \multirow{2}{*}{Method} &
        \multicolumn{2}{c}{\textbf{Knowledge}} &
        \multicolumn{2}{c}{\textbf{Mathematics}} &
        \multicolumn{2}{c}{\textbf{Medicine}} &
        \multicolumn{1}{c}{\textbf{Science}} &
        \multirow{2}{*}{Avg.} \\
        \cmidrule(lr){3-4} \cmidrule(lr){5-6} \cmidrule(lr){7-8} \cmidrule(lr){9-9}
        & & MMLU & MMLU-Pro & MATH & GSM-Hard & MedMCQA & MedQA & GPQA & \\
    \midrule

    \multirow{6}{*}{LLaMA3.1-8B}
    & IO & 71.8 & 46.2 & 48.4 & 34.8 & 59.8 & 65.4 & \textbf{30.6} & 51.0 \\
    & CoT~\cite{wei2022chain} & 71.6 & 46.6 & 47.8 & \textbf{36.2} & \underline{61.0} & \underline{67.2} & 27.0 & 51.1 \\
    & SC~\cite{sc} & \underline{73.4} & \underline{47.6} & \textbf{56.8} & \underline{35.4} & 60.0 & 65.0 & \underline{29.9} & \underline{52.6} \\
    & LLM Debate~\cite{debate1} & \textbf{74.6} & \textbf{53.2} & \underline{55.8} & 34.0 & \textbf{61.4} & \textbf{68.8} & 29.2 & \textbf{53.9} \\
    & Div-MAD~\cite{debate2} & 53.6 & 32.4 & 38.6 & 22.0 & 45.6 & 49.4 & 25.9 & 38.2 \\
    & DMAD~\cite{liu2025breaking} & 67.4 & 35.8 & 34.6 & 21.2 & 54.8 & 59.0 & 25.7 & 42.6 \\
    \midrule

    \multirow{6}{*}{InternLM3-8B}
    & IO & 76.8 & \textbf{58.2} & 74.4 & 51.2 & 60.2 & 63.8 & \underline{37.3} & 60.3 \\
    & CoT~\cite{wei2022chain} & 77.0 & 53.6 & 77.2 & 52.0 & 59.6 & 62.4 & \textbf{37.9} & 59.9 \\
    & SC~\cite{sc} & \underline{78.2} & \underline{57.8} & \textbf{77.6} & \underline{53.6} & \textbf{62.2} & \underline{64.6} & 36.4 & \textbf{61.5} \\
    & LLM Debate~\cite{debate1} & \textbf{78.6} & \underline{57.8} & \underline{77.5} & \textbf{54.8} & 60.4 & \textbf{66.4} & 33.5 & \underline{61.3} \\
    & Div-MAD~\cite{debate2} & 73.0 & 50.4 & 76.4 & \textbf{54.8} & 55.2 & 54.4 & 35.8 & 57.1 \\
    & DMAD~\cite{liu2025breaking} & 76.0 & 55.0 & 75.2 & 53.2 & \underline{60.6} & 62.4 & 35.0 & 59.6 \\
    \midrule

    \multirow{6}{*}{Qwen2.5-7B}
    & IO & 71.8 & 54.2 & 75.0 & 53.6 & 56.2 & \underline{61.4} & \textbf{35.0} & 58.2 \\
    & CoT~\cite{wei2022chain} & 72.0 & 53.8 & 78.4 & \underline{57.0} & 54.8 & 59.2 & 32.8 & 58.3 \\
    & SC~\cite{sc} & 71.8 & \underline{55.4} & \textbf{81.4} & 56.8 & \textbf{56.6} & 60.0 & \underline{34.4} & \underline{59.5} \\
    & LLM Debate~\cite{debate1} & \textbf{73.8} & \textbf{57.4} & \underline{78.6} & \textbf{57.4} & \underline{56.4} & \textbf{63.4} & 33.3 & \textbf{60.0} \\
    & Div-MAD~\cite{debate2} & 70.8 & 50.0 & 40.6 & 24.2 & 49.0 & 53.0 & 29.7 & 45.3 \\
    & DMAD~\cite{liu2025breaking} & 71.6 & 52.0 & 69.0 & 54.6 & 53.8 & 60.4 & 26.6 & 55.4 \\
    \midrule

    \multirow{6}{*}{Qwen2.5-14B}
    & IO & 76.4 & 65.0 & 79.8 & \textbf{61.2} & 63.6 & 68.8 & \textbf{42.2} & 65.3 \\
    & CoT~\cite{wei2022chain} & \textbf{77.8} & 63.8 & 80.8 & \textbf{61.2} & 63.4 & 68.8 & 39.7 & 65.1 \\
    & SC~\cite{sc} & 76.2 & \underline{66.0} & \underline{82.8} & \underline{60.8} & 63.4 & 70.8 & \textbf{42.2} & \underline{66.0} \\
    & LLM Debate~\cite{debate1} & \underline{77.4} & \textbf{67.2} & \textbf{84.2} & \textbf{61.2} & \underline{66.2} & \underline{71.4} & \underline{41.3} & \textbf{67.0} \\
    & Div-MAD~\cite{debate2} & 76.8 & 63.0 & 80.4 & 60.6 & 58.8 & 65.0 & 40.6 & 63.6 \\
    & DMAD~\cite{liu2025breaking} & \textbf{77.8} & 61.6 & 78.6 & 59.6 & \textbf{66.6} & \textbf{74.6} & 35.3 & 64.8 \\
    \midrule

    \multirow{6}{*}{GPT-4o-mini}
    & IO & \underline{82.2} & 60.6 & 75.4 & 56.8 & \textbf{69.8} & 79.2 & \underline{39.5} & 66.2 \\
    & CoT~\cite{wei2022chain} & \textbf{82.4} & 64.8 & 77.6 & 57.4 & \underline{69.2} & 80.4 & 35.7 & 66.8 \\
    & SC~\cite{sc} & 80.8 & \underline{66.4} & \underline{80.6} & \underline{57.6} & 68.6 & 80.4 & \textbf{43.3} & \underline{68.2} \\
    & LLM Debate~\cite{debate1} & 81.6 & \textbf{68.0} & \textbf{81.4} & \textbf{59.4} & \textbf{69.8} & \textbf{82.4} & 38.1 & \textbf{68.7} \\
    & Div-MAD~\cite{debate2} & 80.6 & 55.0 & 76.0 & 53.0 & 59.2 & 70.6 & 33.0 & 60.1 \\
    & DMAD~\cite{liu2025breaking} & 81.8 & 63.6 & 75.4 & 52.3 & 69.1 & \underline{82.0} & 30.3 & 64.9 \\
    \bottomrule
    \end{tabular*}
    \caption{Performance of different methods on unimodal datasets. Best are \textbf{bolded} and second-best are \underline{underlined}.}
    \label{tab:main_unimodal}
\end{table*}

\subsection{Evaluation Metrics} 
We adopt a set of evaluation metrics to assess MAD methods from both performance and efficiency perspectives. 
Accuracy serves as the primary metric for effectiveness. 
To capture practical cost considerations, we also track \textit{token consumption} and \textit{inference time}, explicitly distinguishing between input and output tokens to isolate the costs of accumulated debate history from active agent generation. 
By evaluating both accuracy and fine-grained costs, we can assess whether performance improvements justify the associated computational overhead.

\begin{table*}[t]
    \centering
\small
\setlength{\tabcolsep}{5.2pt}
\renewcommand{\arraystretch}{1.05}
\begin{tabular*}{0.9\textwidth}{@{\extracolsep{\fill}}llccccccc@{}}
    \toprule
        \multirow{2}{*}{Model} & \multirow{2}{*}{Method} &
        \multicolumn{1}{c}{\textbf{Knowledge}} &
        \multicolumn{2}{c}{\textbf{Mathematics}} &
        \multicolumn{1}{c}{\textbf{Medicine}} &
        \multicolumn{2}{c}{\textbf{Reasoning}} &
        \multirow{2}{*}{Avg.} \\
        \cmidrule(lr){3-3} \cmidrule(lr){4-5} \cmidrule(lr){6-6} \cmidrule(lr){7-8}
        & & MME & MathVista & MathVision & PathVQA & MME-Reas. & VisualPuzzles & \\
    \midrule

    \multirow{6}{*}{Qwen2.5-VL-7B}
    & IO & 73.4 & 64.8 & 25.2 & \underline{45.4} & 27.8 & 34.4 & 45.2 \\
    & CoT~\cite{wei2022chain} & 67.8 & 66.6 & 23.0 & 26.0 & 32.0 & 34.8 & 41.7 \\
    & SC~\cite{sc} & \underline{76.8} & \underline{67.0} & 27.8 & 27.0 & 30.0 & \underline{36.8} & 44.2 \\
    & LLM Debate~\cite{debate1} & 72.4 & \textbf{67.8} & \textbf{28.6} & 42.6 & \textbf{33.8} & \textbf{40.2} & \underline{47.6} \\
    & Div-MAD~\cite{debate2} & 57.8 & 64.2 & 26.2 & 25.6 & 26.4 & 33.8 & 39.0 \\
    & DMAD~\cite{liu2025breaking} & \textbf{78.2} & 65.6 & \underline{28.4} & \textbf{46.2} & \underline{32.8} & 36.6 & \textbf{48.0} \\
    \midrule

    \multirow{6}{*}{LLaVA-Next-7B}
    & IO & 40.2 & 24.0 & 8.0 & 11.2 & 12.8 & 18.0 & 19.0 \\
    & CoT~\cite{wei2022chain} & 30.2 & 24.0 & 7.2 & 4.6 & \underline{14.8} & 16.0 & 16.1 \\
    & SC~\cite{sc} & 29.2 & 24.8 & 9.8 & 3.4 & 13.4 & \underline{20.0} & 16.8 \\
    & LLM Debate~\cite{debate1} & 45.0 & \textbf{28.8} & \underline{10.2} & 13.2 & \textbf{16.2} & 18.6 & \underline{22.0} \\
    & Div-MAD~\cite{debate2} & \underline{50.8} & 20.7 & 8.0 & \underline{18.0} & 13.0 & 18.0 & 21.4 \\
    & DMAD~\cite{liu2025breaking} & \textbf{54.0} & \underline{27.4} & \textbf{11.6} & \textbf{19.4} & \underline{14.8} & \textbf{22.8} & \textbf{25.0} \\
    \midrule

    \multirow{6}{*}{InternVL3-8B}
    & IO & 85.0 & 65.8 & 27.0 & \textbf{52.0} & 30.4 & 38.2 & 49.7 \\
    & CoT~\cite{wei2022chain} & 85.0 & \underline{66.6} & 27.6 & 44.4 & 29.0 & 34.8 & 47.9 \\
    & SC~\cite{sc} & \textbf{86.0} & 65.0 & 29.0 & 32.0 & 28.8 & 37.8 & 46.4 \\
    & LLM Debate~\cite{debate1} & \underline{85.2} & \underline{66.6} & 28.0 & \textbf{52.0} & \textbf{33.0} & \underline{40.8} & \underline{50.9} \\
    & Div-MAD~\cite{debate2} & 50.2 & 50.4 & \underline{29.2} & 38.4 & 28.8 & 37.4 & 39.1 \\
    & DMAD~\cite{liu2025breaking} & \underline{85.2} & \textbf{67.2} & \textbf{30.6} & \underline{50.8} & \underline{31.0} & \textbf{41.8} & \textbf{51.1} \\
    \midrule

    \multirow{6}{*}{InternVL3-14B}
    & IO & 76.8 & 66.8 & 30.0 & \underline{49.2} & 33.2 & 40.2 & 49.4 \\
    & CoT~\cite{wei2022chain} & 76.2 & 67.8 & 31.6 & 34.8 & 31.6 & 41.0 & 46.5 \\
    & SC~\cite{sc} & \underline{82.6} & 68.8 & 35.2 & 39.0 & 34.2 & 43.2 & 50.5 \\
    & LLM Debate~\cite{debate1} & 82.2 & \textbf{71.8} & 34.6 & 43.4 & \textbf{38.4} & \textbf{54.2} & \underline{52.6} \\
    & Div-MAD~\cite{debate2} & 66.4 & \underline{69.2} & \textbf{37.2} & 43.2 & \underline{36.6} & 46.2 & 49.8 \\
    & DMAD~\cite{liu2025breaking} & \textbf{86.4} & \underline{69.2} & \underline{35.8} & \textbf{54.8} & 35.0 & \underline{47.6} & \textbf{54.8} \\
    \midrule

    \multirow{6}{*}{GPT-4o-mini}
    & IO & \underline{72.6} & 57.0 & \textbf{32.0} & \textbf{39.6} & 26.0 & 38.8 & \underline{44.3} \\
    & CoT~\cite{wei2022chain} & 65.2 & 57.6 & 26.2 & 14.4 & \textbf{30.0} & 37.4 & 38.5 \\
    & SC~\cite{sc} & 63.2 & \underline{59.4} & \underline{30.8} & 18.0 & 27.4 & \underline{40.6} & 39.9 \\
    & LLM Debate~\cite{debate1} & 71.2 & \textbf{59.6} & 28.8 & 29.4 & \underline{29.8} & \textbf{43.4} & 43.7 \\
    & Div-MAD~\cite{debate2} & 61.8 & 51.4 & 28.0 & 21.0 & 27.6 & 37.8 & 37.9 \\
    & DMAD~\cite{liu2025breaking} & \textbf{76.0} & 56.2 & 27.4 & \underline{37.5} & 28.9 & 40.4 & \textbf{44.4} \\
    \bottomrule
    \end{tabular*}
    \caption{Performance of different methods on multimodal datasets. Best are \textbf{bolded} and second-best are \underline{underlined}.}
    \label{tab:main_multimodal}
\end{table*}

\section{Experiments}

\vspace{2mm}

\subsection{Settings}

\vspace{2mm}

\noindent \textbf{Base model.} To comprehensively evaluate existing MAD methods, we select six open-source models spanning both unimodal and multimodal types: LLaMA3.1-8B~\cite{llama3}, Qwen2.5-7B~\cite{qwen25}, InternLM3-8B~\cite{internlm2}, Qwen2.5VL-7B~\cite{qwen25vl}, LLaVA-Next-7B~\cite{llava-1.5} and InternVL3-8B~\cite{internvl3}. 
To analyze the impact of model size on performance, we further conduct experiments with larger-scale models, namely Qwen2.5-14B~\cite{qwen25} and InternVL-14B~\cite{internvl3}.
Moreover, we include the closed-source model GPT-4o-mini~\cite{gpt-4o-mini} in our evaluation.
These models cover a wide range of architectures and modalities.
Importantly, each MAD method is evaluated across all models, ensuring a consistent and comprehensive assessment of performance across architectures, scales, and modalities.

\vspace{2mm}

\noindent \textbf{Implementation Details.} Following MASLab~\cite{maslab} and for cost efficiency, we randomly sample 500 examples from each dataset for evaluation. 
We set the temperature to 0.5 and the maximum token length to 4096. 
To achieve more precise answer assessment, we use a model specifically designed for evaluation, x-Verify~\cite{xVerify}, to determine the final answers, adopting an LLM-as-a-judge paradigm to ensure scalable and consistent evaluation across diverse tasks and modalities.

\vspace{2mm}

\textbf{For the evaluated methods, we keep all method-specific hyperparameter settings (e.g., the number of agents and debate rounds) consistent with their official implementations. }
Specifically, Self-Consistency (SC) uses $N=5$ independent reasoning paths with majority voting. LLM Debate uses $N=3$ agents and $T=2$ debate rounds. Div-MAD uses two role-based agents (``Affirmative'' and ``Negative'') with $T=3$ rounds. DMAD uses $N=3$ agents and $T=3$ rounds. 
Following the original paper, DMAD assigns different reasoning strategies to agents to encourage diversity. For unimodal tasks, we use Chain-of-Thought (CoT) \cite{wei2022chain}, Step-Back Prompting (SBP) \cite{zheng2024take}, and Program-of-Thought (PoT) \cite{chen2023program}.
For multimodal tasks, we use IO inference, Compositional Chain-of-Thought (CCoT) \cite{mitra2024compositional}, and Duty-Distinct Chain-of-Thought (DDCoT) \cite{zheng2023ddcot}. 
Experiments are conducted on eight NVIDIA 40G A100 GPUs, while GPT-4o-mini is accessed via OpenAI API calls.

\subsection{Main Results}

Tables~\ref{tab:main_unimodal} and~\ref{tab:main_multimodal} summarize the results across all unimodal and multimodal datasets, from which we draw the following conclusions:

\begin{greybox}
\raisebox{-0.15\height}{\includegraphics[height=12pt]{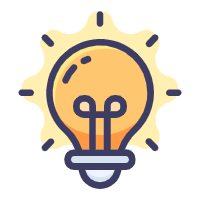}}~\textbf{Insight 1}: MAD methods show mixed effectiveness: some approaches can improve over single-agent baselines, while others may underperform.
\end{greybox}
\noindent Specifically, LLM Debate improves performance over single-agent baselines (IO, CoT) on the majority of datasets and base models in both unimodal and multimodal settings.
In contrast, other MAD variants, including Div-MAD and DMAD with weaker base models like LLaMA3.1-8B, often underperform, sometimes even falling below single-agent baselines (e.g., Div-MAD averages 38.2 vs. IO's 51.0 on LLaMA3.1-8B). The results show that the MAD could be harmful than helpful, which aligns with previous findings~\cite{wynn2025talk}.

\begin{greybox}
\raisebox{-0.15\height}{\includegraphics[height=12pt]{figs/light.png}}~\textbf{Insight 2}: Collaborative debate consistently outperforms adversarial paradigms in both performance and robustness across all modalities.
\end{greybox}
\noindent In unimodal settings, LLM Debate secures the highest average accuracy across the majority of base models. 
DMAD consistently outperforms the adversarial Div-MAD by a substantial margin.
This disparity is further amplified in multimodal settings, where DMAD emerges as a top performer while Div-MAD continues to exhibit subpar results.
We argue that adversarial paradigms often introduce "divergent noise" and conflicting logic that can derail the reasoning chain. 
In contrast, collaborative frameworks foster a synergy that acts as a collective self-correction mechanism, effectively filtering out unimodal hallucinations and reconciling visual-linguistic inconsistencies more reliably than friction-based approaches.

\begin{figure*}[t]
\includegraphics[width=1\linewidth]{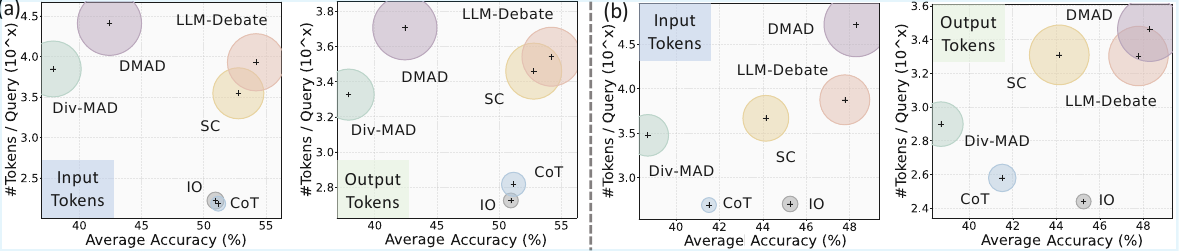} 
  \caption {Accuracy vs. token consumption (\#Tokens/Query) on log-scale. Results are shown for (a) LLaMA3.1-8B on unimodal datasets and (b) Qwen2.5-VL-7B on multimodal datasets. Left and right plots correspond to Input and Output tokens, respectively.
  }
  \label{fig:acctoken}
\end{figure*}

\begin{greybox}
\raisebox{-0.15\height}{\includegraphics[height=12pt]{figs/light.png}}~\textbf{Insight 3}: The effectiveness of MAD varies across tasks, with more pronounced improvements on logic- and reasoning-oriented benchmarks than on knowledge-focused ones.
\end{greybox}
\noindent For instance, using Qwen2.5-14B as the base model, LLM Debate improves performance on MATH from 79.8 to 84.2, while on MMLU the improvement is marginal, increasing only from 64.0 to 65.0. 
Similarly, using InternVL3-14B as the base model, LLM Debate boosts VisualPuzzles performance from 40.2 to 54.2.
This pattern suggests that MAD is particularly effective when tasks require systematic reasoning or complex problem-solving, whereas tasks relying primarily on factual recall benefit less from such interaction.

\begin{greybox}
\raisebox{-0.15\height}{\includegraphics[height=12pt]{figs/light.png}}~\textbf{Insight 4}: Introducing multiple perspectives in debate improves multimodal performance but offers little or negative gains in unimodal tasks.
\end{greybox}

\noindent For unimodal tasks, LLM Debate's standard multi-turn consensus is generally sufficient to fully exploit the limited input information. 
Introducing multiple perspectives, as in DMAD, provides almost no improvement and can even harm performance, as the sparse unimodal inputs offer few complementary signals and conflicting perspectives may mislead the reasoning process.
In contrast, multimodal tasks contain rich, complementary information across modalities, such as visual and textual cues. 
Here, multi-perspective debate enables DMAD to integrate and reconcile these signals, substantially enhancing performance. 
This contrast highlights that multi-perspective consensus mechanisms are more effective in information-rich multimodal settings, whereas their potential is constrained in unimodal scenarios.

\begin{greybox}
\raisebox{-0.15\height}{\includegraphics[height=12pt]{figs/light.png}}~\textbf{Insight 5}: MAD effectiveness is affected by the performance of the underlying base model.
\end{greybox}
\noindent Generally, for collaborative debate methods, stronger base models lead to better MAD performance. 
A similar trend holds for adversarial approaches; however, the difference between Div-MAD and IO is substantial when the base model is weak, whereas this gap narrows as the base model's capabilities improve. 
We argue that weaker models amplify the negative impact of adversarial interactions, introducing more divergent or misleading reasoning paths, while stronger models are better able to resist such noise and maintain more consistent performance.

\subsection{Efficiency Result} 

\begin{figure}[t]
\centering
\includegraphics[width=\columnwidth]{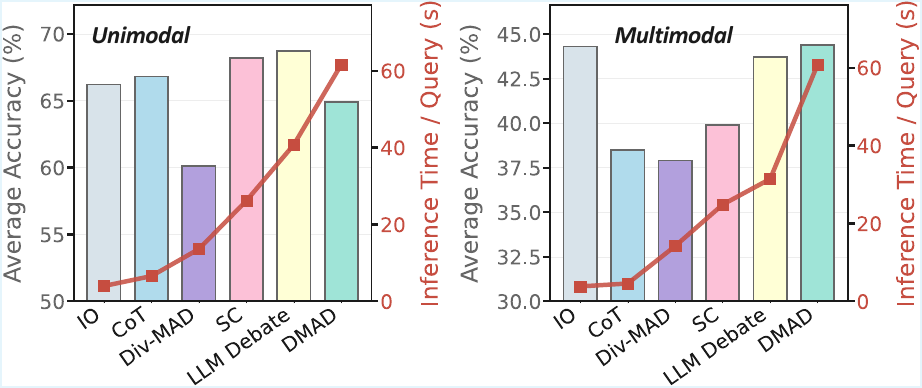}
    \caption{Accuracy vs. Inference Time (GPT-4o-mini). Bars denote accuracy (left) and lines denote time cost (right) across (a) unimodal and (b) multimodal datasets.}
    \label{fig:inference_time}
    \vspace{-3mm}
\end{figure}

Figure~\ref{fig:acctoken} illustrates these results of the trade-off between token usage and model accuracy for LLaMA3.1-8B and Qwen2.5-VL-7B across unimodal and multimodal tasks, respectively. 
A substantial efficiency gap is evident, as MAD methods like DMAD and LLM-Debate consume significantly more tokens than single-agent baselines while bringing little improvement. 
This surge stems primarily from the extensive context needed to maintain debate history (Input Tokens) and the cumulative responses generated by multiple agents (Output Tokens).
In addition to token consumption, we evaluate real-world inference time. To minimize measurement variance from local GPU hardware fluctuations, we utilize the stable API of GPT-4o-mini for this analysis. As shown in Figure~\ref{fig:inference_time}, the trend mirrors our token findings, where MAD methods take considerably longer to complete queries compared to standard approaches.

\begin{greybox}
\raisebox{-0.15\height}{\includegraphics[height=12pt]{figs/light.png}}~\textbf{Insight 6}: MAD methods incur high token and time costs with limited performance gains.
\end{greybox}

\section{Analysis and Discussion}

\subsection{Effect of Model Heterogeneity}

\noindent\textbf{Do heterogeneous models boost performance?} Previous MAD methods typically employ the same model for all debating agents and increase the decoding temperature to enhance diversity. 
Beyond stochastic sampling, an alternative strategy to improve agent diversity is assigning heterogeneous models to different agents. 
To investigate whether such model heterogeneity further improves performance, we conduct an analytical experiment. 
Three agents are instantiated using a pool of 7B--8B models listed in Table~\ref{tab:main_unimodal} and Table~\ref{tab:main_multimodal}. Table~\ref{tab:combined_diversity_avg_singlecol} presents the comparative results. 
In this analysis, Worst Homog. and Best Homog. represent the worst and best performance achieved, respectively, when all agents use the same model. 
Our empirical results show that while the performance of heterogeneous models occasionally exceeds the Best Homog. baseline, most results fall within the range between the worst and best homogeneous configurations. 
This indicates that simply employing heterogeneous models for agents does not yield a substantial performance leap.

\begin{table}[ht]
\centering
\small
\renewcommand{\arraystretch}{1} 
\setlength{\tabcolsep}{0pt} 
\begin{tabular*}{\columnwidth}{@{\extracolsep{\fill}} lccc}
\toprule
\textbf{Methods} & \textbf{Worst Homog.} & \textbf{Heterogeneous} & \textbf{Best Homog.} \\
\midrule
\multicolumn{4}{l}{\textit{Unimodal Datasets (Average)}} \\
\midrule
LLM Debate & 63.6 & \textbf{72.9} & \textbf{72.9} \\
Div-MAD    & 45.9 & 62.4 & \textbf{66.2} \\
DMAD       & 54.1 & 66.3 & \textbf{67.0} \\
\midrule
\multicolumn{4}{l}{\textit{Multimodal Datasets (Average)}} \\
\midrule
LLM Debate & 29.3 & 61.9 & \textbf{62.6} \\
Div-MAD    & 27.3 & \textbf{49.6} & 46.0 \\
DMAD       & 26.3 & 58.3 & \textbf{63.2} \\
\bottomrule
\end{tabular*}
\caption{Average performance comparison between heterogeneous models and homogeneous configurations. The Unimodal average includes MATH, MMLU, and MedMCQA; the Multimodal average includes MME, MME-Reas., and MathVista.
Best results are \textbf{bolded}. }
\label{tab:combined_diversity_avg_singlecol}
\end{table}

Combined with our previous \textbf{Insight 4}, although diversity of perspectives can drive performance, model heterogeneity alone does not guarantee it. Variance among models does not necessarily produce distinct reasoning paths or complementary problem-solving behaviors, so simply mixing models may fail to outperform the best individual configuration.

\begin{greybox}
\raisebox{-0.15\height}{\includegraphics[height=12pt]{figs/light.png}}~\textbf{Insight 7}: The inherent differences among heterogeneous models do not necessarily translate into distinct reasoning paths, and in most cases, do not lead to performance improvements.
\end{greybox}

\subsection{Scaling Analysis}

\begin{figure}[t]
\centering
\includegraphics[width=\columnwidth]{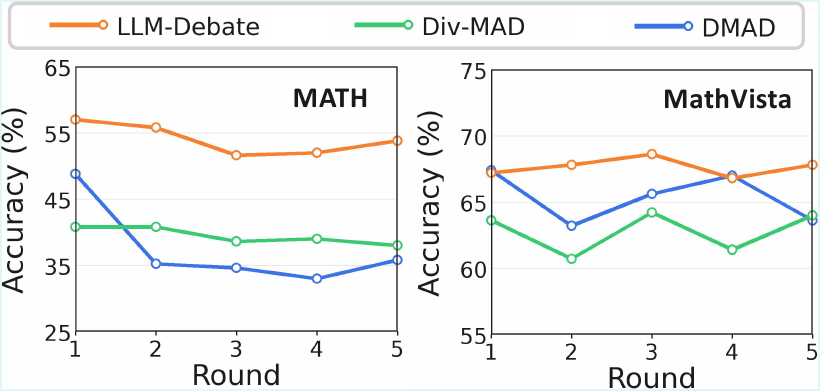}
    \caption{Performance across debate rounds. Evaluated on (a) MATH (LLaMA3.1-8B) and (b) MathVista (Qwen2.5-VL-7B).}
    \label{fig:round}
\end{figure}

\noindent\textbf{Does increasing debate rounds help?} 
Intuitively, one might expect that extending the debate would continuously refine reasoning and improve accuracy. 
However, as shown in Figure~\ref{fig:round}, we observe that performance tends to fluctuate or plateau rather than strictly improve across rounds. For instance, on the MATH dataset, accuracy even exhibits a slight downward trend after the initial rounds. 
This phenomenon aligns with recent findings by~\citet{choidebate}, suggesting that without a reliable external signal or intrinsic capability to recognize the truth, extended interaction does not guarantee convergence to a better answer.
Instead, the debate process often merely "shuffles" opinions: while some errors may be corrected, valid arguments can also be subverted by misleading consensus, resulting in relatively flat overall performance.

\begin{table}[h]
\centering
\small
\begin{tabular}{@{}l|ccccc@{}}
\toprule
Agent Number & 2 & 3 & 4 & 5 & 6 \\ 
\midrule
MATH & 53.4 & 55.8 & 53.4 & \textbf{57.0} & 56.6  \\ 
MathVista & 66.6 & 67.8 & 67.2 & 66.4 & \textbf{70.2}   \\ 
\bottomrule
\end{tabular}
\caption{Impact of agent quantity on LLM-Debate. Results on MATH (LLaMA3.1-8B) and MathVista (Qwen2.5-VL-7B). Best results are \textbf{bolded}.}
\label{tab:agent_num}
\end{table}

\noindent\textbf{Do more agents lead to better results?} 
We conduct analysis on LLM-Debate, as Div-MAD typically requires a fixed number of agents, and DMAD relies on distinct inference paths that are difficult to scale arbitrarily. Table~\ref{tab:agent_num} presents the performance of LLM-Debate as the number of agents grows from 2 to 6.
In contrast to increasing rounds, increasing the population size generally yields positive gains. 
 We observe an overall upward trend in accuracy for both MATH (53.4 to 56.6) and MathVista (66.6 to 70.2). 
This improvement can be attributed to the "ensemble effect". 
With a larger population, the diversity of knowledge increases, raising the probability that at least one agent holds the correct reasoning path to influence the group. 
Consequently, expanding the agent scale proves to be a more reliable strategy for improvement than simply prolonging the debate duration.

\begin{greybox}
\raisebox{-0.15\height}{\includegraphics[height=12pt]{figs/light.png}}~\textbf{Insight 8}: More debate rounds offer little benefit, while more agents generally improve performance.
\end{greybox}

\subsection{Error Attribution}
To analyze failure modes in MAD, we uniformly sample 100 failed cases for manual error attribution. 
Similar to the taxonomy in previous work~\cite{cemri2025why,wynn2025talk}, we group errors into four categories:
(1) \textit{Incorrect Conformity}, where a correct agent conforms to an incorrect consensus due to confidence bias or peer pressure;
(2) \textit{Collective Delusion}, where agents mutually reinforce incorrect assumptions, forming a hallucination loop;
(3) \textit{Context Degradation}, where debate quality deteriorates over rounds due to repetition or format drift;
and (4) \textit{Selection Failure}, where a correct answer appears during the debate but is missed by the final aggregation mechanism.

\begin{figure}[t]
\centering
\includegraphics[width=0.89\linewidth]{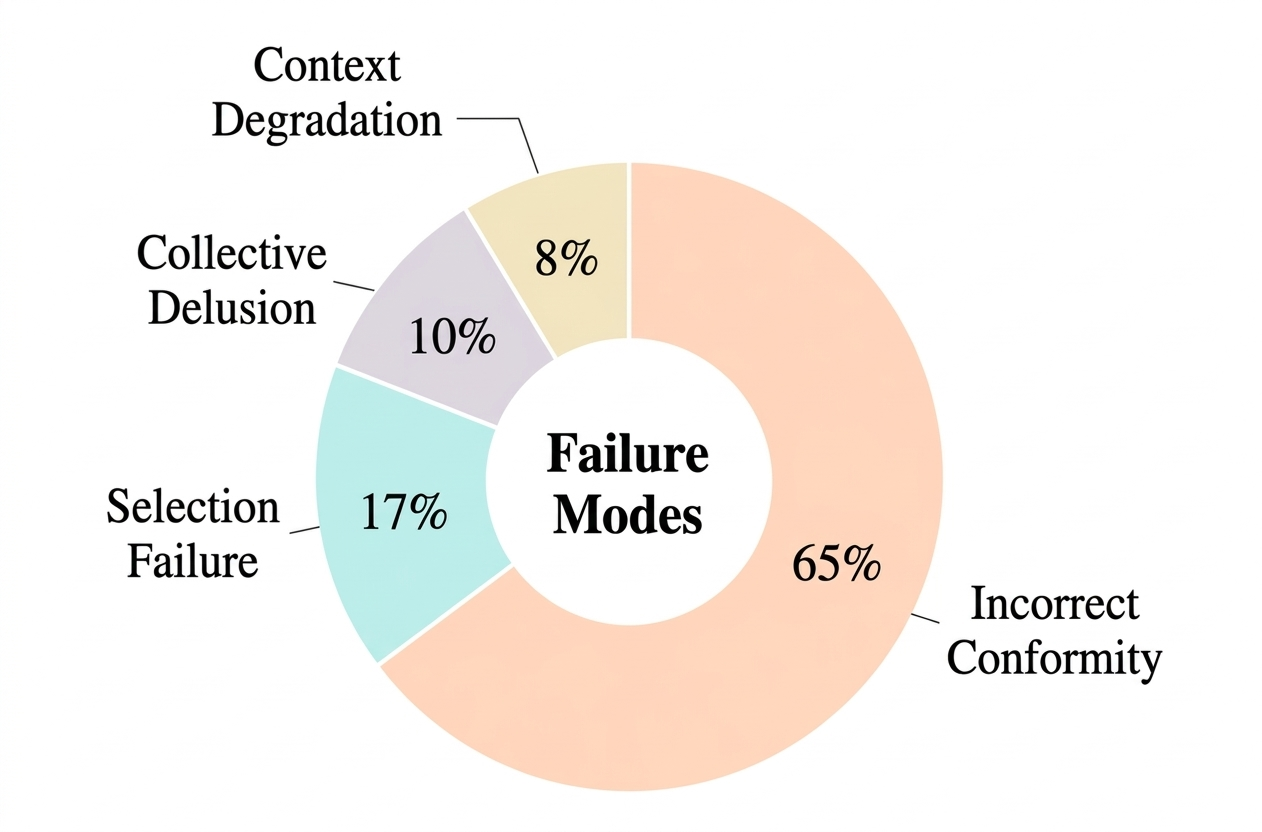} 
    \caption {Distribution of failure modes in MAD.}
      \label{fig:attribute}
      \vspace{-3mm}
\end{figure}

As shown in Figure~\ref{fig:attribute}, \textit{Collective Delusion} accounts for the majority (65\%) of errors, revealing that agent interaction often reinforces incorrect claims rather than correcting them. 
Additionally, \textit{Selection Failure} (17\%) indicates that correct answers often emerge during debate but are ultimately missed by the aggregator. 
These findings suggest that future work should prioritize mitigating groupthink and improving answer selection.

\begin{greybox} \raisebox{-0.15\height}{\includegraphics[height=12pt]{figs/light.png}}~\textbf{Insight 9}: Agents tend to reinforce mutual errors rather than self-correcting, and aggregators often fail to identify valid reasoning already present in the debate history. \end{greybox}

\subsection{Case Study} 
\begin{figure}[t]
\centering
\includegraphics[width=1\linewidth]{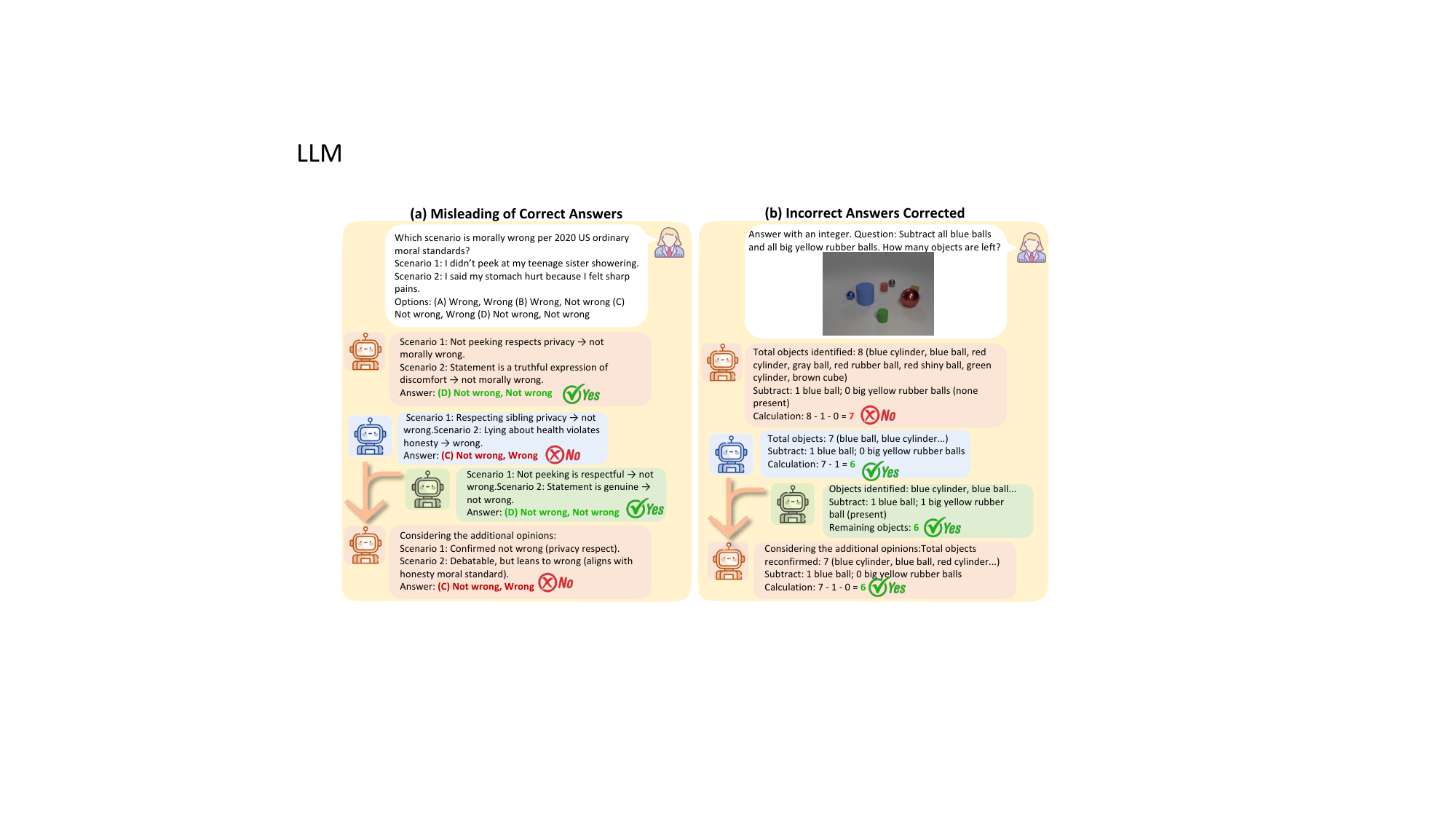} 
  \caption {Case study of LLM Debate: (a) Misleading of Correct Answers and (b) Incorrect Answers Corrected.}
  \label{fig:case_llm_debate}
\end{figure}

To further explore the internal interaction dynamics of MAD, we analyze the debate history and find two representative cases, as shown in Figure~\ref{fig:case_llm_debate}. 
These examples provide an intuitive view of how multi-agent discussion can either help or hinder the final decision-making process.
The first case, \textit{misleading of correct answers}, shows an agent being swayed away from an initially correct line of reasoning by incorrect peer responses, eventually leading to a group-level ``herding effect.'' In contrast, the second case, \textit{correction of incorrect answers}, illustrates how debate can help agents identify and rectify logical flaws in each other's reasoning, thereby mitigating individual hallucinations and improving the final outcome. 
These qualitative findings reveal that while MAD can leverage collective intelligence to refine reasoning, its success depends heavily on the agents' ability to balance consensus with independent verification and resist being misled by persuasive but incorrect arguments.

\section{Conclusion}

In this paper, we address the critical gaps of fragmentation and unimodal constraints in existing MAD studies by introducing M3MAD-Bench, a comprehensive benchmark characterized by multi-domain tasks, multi-modal inputs, and multi-metric evaluations. 
We conduct an extensive assessment of MAD methodologies across nine diverse models and systematically analyze how pivotal factors, including architectural diversity (homogeneity vs. heterogeneity), agent population, and debate rounds, influence collective performance. 
By performing a rigorous error attribution, we identify collective delusion as the primary socio-cognitive pathologies leading to debate failures.
Future research may consider exploring intervention mechanisms to mitigate collective delusion and designing adaptive protocols to balance multimodal information richness with multi-perspective reasoning diversity.
We believe M3MAD-Bench provides a solid empirical foundation offers guidance for future research on multi-agent collective intelligence.

\section*{Acknowledgement}
The project is supported by the Shandong Provincial Natural Science Foundation (ZR2025QC1570, 	
ZR2026LLX017), and CAAI-Lenovo Blue Sky Research Fund
(2025CAAI-LENOVO-11).

\section*{Ethics and Privacy Statement}
This work primarily focuses on the evaluation of multi-agent debate methods using publicly available benchmarks and does not involve the collection of new personal data or human-subject studies. We anticipate minimal societal risk, while recognizing that future applications of multi-agent systems should continue to consider responsible deployment, fairness, and transparency.

\bibliographystyle{ACM-Reference-Format}
\bibliography{sample-base}

\end{document}